\documentclass[11pt,a4paper]{article}
\usepackage[hyperref]{emnlp2020}
\usepackage{times}
\usepackage{latexsym}

\usepackage{microtype}

\aclfinalcopy 
\def\aclpaperid{1675} 

\usepackage{amssymb}

\usepackage{latexsym}

\usepackage{url}

\usepackage{multirow}
\usepackage{balance}
\usepackage{amsfonts}
\usepackage{amsmath}
\usepackage{bbold}

\usepackage{amsthm}
\usepackage{graphicx}

\usepackage{microtype}
\usepackage{url}
\usepackage{enumitem}
\usepackage{wrapfig,lipsum}
\usepackage{multirow}
\usepackage{makecell}
\usepackage{wrapfig}
\usepackage{tabularx}
\usepackage[boxed, ruled,vlined,linesnumbered]{algorithm2e}

\usepackage{subcaption}
\usepackage{float}
\usepackage{booktabs}
\usepackage{caption}
\usepackage{blindtext}
\usepackage{amsmath}
\usepackage[linesnumbered,ruled]{algorithm2e}%

\usepackage{pifont}
\usepackage{tabularx,ragged2e}

\usepackage{environ}
\newcommand{\ie}{\emph{i.e.}}

\newcommand{\bs}[1]{\boldsymbol{#1}}
\newcommand{\lb}{\mbox{$\langle$}}
\newcommand{\rb}{\mbox{$\rangle$}}
\newcommand{\jasen}{\textbf{JASen}\xspace}

\definecolor{RoseQuartzBg}{HTML}{F7CAC9}
\definecolor{RoseQuartz}{HTML}{F5A798}
\definecolor{Serenity}{HTML}{92A8D1}
\definecolor{OrangeRed}{rgb}{1.0, 0.27, 0.0}
\definecolor{Turquoise}{HTML}{0F4C81}
\usepackage{xparse}
\NewDocumentCommand{\heng}{ mO{} }{\textcolor{OrangeRed}{\textsuperscript{\textit{Heng}}\textsf{\textbf{\small[#1]}}}}

\title{Weakly-Supervised Aspect-Based Sentiment Analysis via Joint Aspect-Sentiment Topic Embedding 
}

\author{Jiaxin Huang, Yu Meng, Fang Guo, Heng Ji, Jiawei Han \\
  University of Illinois at Urbana-Champaign, IL, USA \\
  \texttt{\{jiaxinh3, yumeng5, fangguo1, hengji, hanj\}@illinois.edu} \\ 
  }

\date{}

\begin{document}
\maketitle
\begin{abstract}
Aspect-based sentiment analysis of review texts is of great value for understanding user feedback in a fine-grained manner. 
It has in general two sub-tasks: (i) extracting aspects from each review, and (ii) classifying aspect-based reviews by sentiment polarity.
In this paper, we propose a weakly-supervised approach for aspect-based sentiment analysis, which uses only a few keywords describing each aspect/sentiment without using any labeled examples.
Existing methods are either designed only for one of the sub-tasks, 
neglecting the benefit of coupling both,
or are based on topic models that may contain overlapping concepts.
We propose to first learn $\lb$sentiment, aspect$\rb$ joint topic embeddings in the word embedding space by imposing regularizations to encourage topic distinctiveness, and then use neural models to generalize the word-level discriminative information by pre-training the classifiers with embedding-based predictions and self-training them on unlabeled data. 
Our comprehensive performance analysis shows that our method generates quality joint topics and outperforms the baselines significantly (7.4\% and 5.1\% F1-score gain on average for aspect and sentiment classification respectively) on benchmark datasets\footnote{Our code and data are available at \url{https://github.com/teapot123/JASen}.}.

\end{abstract}

\section{Introduction}

With the vast amount of reviews emerging on platforms like Amazon and Yelp, aspect-based sentiment analysis, which extracts opinions about certain facets of entities from text, becomes increasingly essential and benefits a wide range of downstream applications~\cite{Bauman2017AspectBR, Nguyen2015SentimentAO}. 

Aspect-based sentiment analysis contains two sub-tasks: Aspect extraction and sentiment polarity classification. 
The former identifies the aspect covered in the review, whereas the latter decides its sentiment polarity.

\begin{figure}[t]
\centering
\includegraphics[width=1\linewidth]{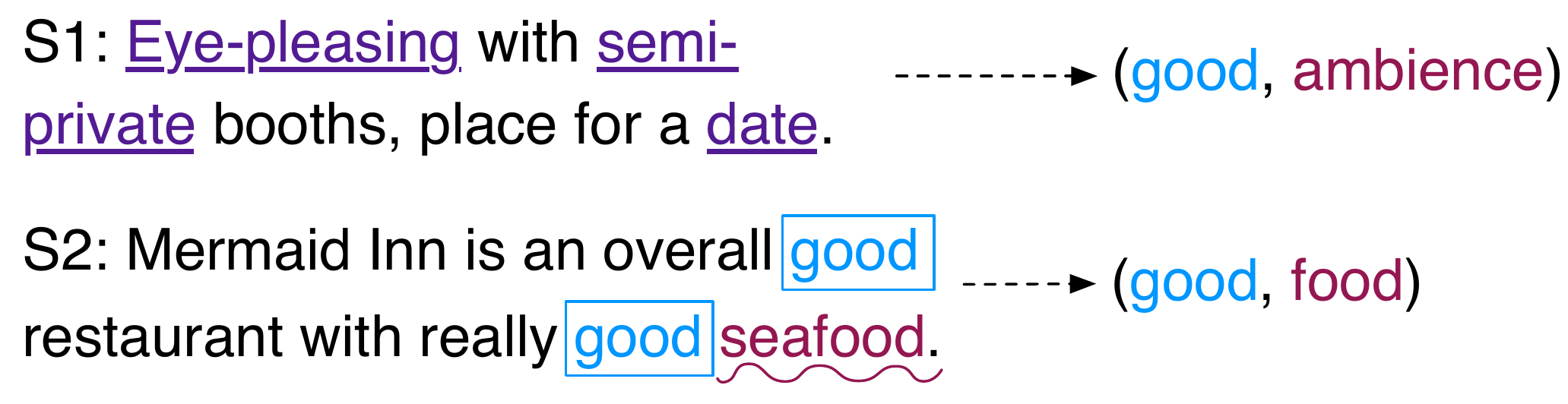}
\caption{
Two sample restaurant reviews. Pure aspect words are in red and wavy-underlined, and general opinion words are in blue and framed in boxes. Words implying both aspects and opinions (which we define as joint topics) are underlined and in purple.
}
\label{fig:exp}
\vspace{-0.5cm}
\end{figure}

Various methods have been proposed for the task.
Neural network models~\cite{Liu2015FinegrainedOM, Xu2018DoubleEA} have outperformed rule-based models~\cite{Hu2004MiningAS, Zhuang2006MovieRM}, but they require large-scale fine-grained labeled data to train, which can be difficult to obtain.
Some other studies leverage word embeddings to solve the aspect extraction problem in an unsupervised~\cite{He2017AnUN, Liao2019CouplingGA} or weakly-supervised setting~\cite{Angelidis2018SummarizingOA, Karamanolakis2019LeveragingJA}, \textit{without} using any annotated documents.
In this work, we study the weakly-supervised setting, where only a few keywords are provided for each aspect and sentiment.

We show two sample restaurant reviews in Fig.~\ref{fig:exp} together with their expected output---aspect and sentiment labels. 
With a closer look at these two example reviews, we observe that S2 includes a general opinion word ``good'' and a pure aspect word ``seafood'', which are separate hints for sentiment and aspect classification respectively. 
S1, on the other hand, does not address the target with plain and general words, but instead use more specific words like ``semi-private'' and ``date'' which are uniquely used when people feel good about the ambience instead of other aspects.
Humans can interpret these unique and fine-grained terms as hints for a joint topic of $\lb$good, ambience$\rb$, but this is hard for models that are solely trained for one sub-task.
If a model can automatically learn the semantics of each joint topic of $\lb$sentiment, aspect$\rb$, it will be able to identify representative terms of the joint topics such as  ``semi-private'' which provide information for aspect and sentiment simultaneously,
and will consequently benefit both aspect extraction and sentiment classification.
Therefore, leveraging more fine-grained information by coupling the two subtasks will enhance both.

Several LDA-based methods consider learning joint topics~\cite{Zhao2010JointlyMA, Wang2015SentimentAspectEB, Xu2012TowardsJE}, but they rely on external resources such as part-of-speech (POS) tagging or opinion word lexicons. A recent LDA-based model~\cite{GarcaPablos2018W2VLDAAU} uses pre-trained word embedding to bias the prior in topic models to jointly model aspect words and opinion words. Though working fairly well, topic models are generative models and do not enforce topic distinctiveness---topic-word distribution can largely overlap among different topics, allowing topics to resemble each other. Besides, topic models yield unstable results, causing large variance in classification results.

We propose the \jasen model for \textbf{J}oint \textbf{A}spect-\textbf{Sen}timent Topic Embedding.
Our general idea is to learn a joint topic representation for each $\lb$sentiment, aspect$\rb$ pair in the shared embedding space with words so that the surrounding words of topic embeddings nicely describe the semantics of a joint topic.
This is accomplished by training topic embeddings and word embeddings on in-domain corpora and modeling the joint distribution of user-given keywords on all the joint topics.
After learning the joint topic vectors, embedding-based predictions can be derived for any unlabeled review.
However, these predictions are sub-optimal for sentiment analysis where word order plays an important role.
To leverage the expressive power of neural models, we distill the knowledge from embedding-based predictions to convolutional neural networks (CNNs)~\cite{Krizhevsky2012ImageNetCW} 
which perform compositional operations upon local sequences. 
A self-training process is then conducted to refine CNNs by using their high-confident predictions on unlabeled data.

We demonstrate the effectiveness of \jasen by conducting experiments on two benchmark datasets and show that our model outperforms all the baseline methods by a large margin. We also show that our model is able to describe joint topics with coherent term clusters.

Our contributions can be summarized as follows: 
(1) We propose a weakly-supervised method \jasen to enhance two sub-tasks of aspect-based sentiment analysis. Our method does \textit{not} need any annotated data but only a few keywords for each aspect/sentiment.
(2) We introduce an embedding learning objective that is able to capture the semantics of fine-grained joint topics of $\lb$sentiment, aspect$\rb$ in the word embedding space. The embedding-based prediction is effectively leveraged by neural models to generalize on unlabeled data via self-training.
(3) We demonstrate that \jasen generates high-quality joint topics and outperforms baselines significantly on two benchmark datasets.  


\section{Related Work}
The problem of aspect-based sentiment analysis can be decomposed into two sub-tasks: aspect extraction and sentiment polarity classification. Most previous studies deal with them individually. There are various related efforts on aspect extraction \cite{He2017AnUN}, which can be followed by  sentiment classification models \cite{He2018ExploitingDK}. Other methods \cite{GarcaPablos2018W2VLDAAU} jointly solve these two sub-tasks by first separating target words from opinion words and then learning joint topic distributions over words.
Below we first review relevant work on aspect extraction (Sec \ref{sec:rel_AE}) and then turn to studies that jointly extract aspects and sentiment polarity (Sec \ref{sec:rel_Joint}).

\subsection{Aspect Extraction}\label{sec:rel_AE}
Early studies towards aspect extraction are mainly based on manually defined rules \cite{Hu2004MiningAS, Zhuang2006MovieRM}, which have been outperformed by supervised neural approaches that do not need labor-intensive feature engineering. While CNN \cite{Xu2018DoubleEA} and RNN \cite{Liu2015FinegrainedOM} based models have shown the powerful expressiveness of neural models, they can easily consume thousands of labeled documents thus suffer from the label scarcity bottleneck.


Various unsupervised approaches are proposed to model different aspects automatically. LDA-based methods \cite{Brody2010AnUA,Chen2014AspectEW} model each document as a mixture of aspects (topics) and output a word distribution for each aspect. Recently, neural models have shown to extract more coherent topics. ABAE~\cite{He2017AnUN} uses an autoencoder to reconstruct sentences through aspect embedding and removes irrelevant words through attention mechanisms.
CAt~\cite{Tulkens2020EmbarrassinglySU} introduces a single head attention calculated by a Radial Basis Function (RBF) kernel to be the sentence summary. The unsupervised nature of these algorithms is hindered by the fact that the learned aspects often do not well align with user's interested aspects, and additional human effort is needed to map topics to certain aspects, not to mention some topics are irrelevant of interested aspects. 

Several weakly-supervised methods address this problem by using a few keywords per aspect as supervision to guide the learning process. MATE~\cite{Angelidis2018SummarizingOA} extends ABAE by initializing aspect embedding using weighted average of keyword embeddings from each aspect. ISWD~\cite{Karamanolakis2019LeveragingJA} co-trains a bag-of-word classifier and an embedding-based neural classifier to generalize the keyword supervision. Other text classification methods leverage pre-trained language model \cite{Meng2020Lotclass} to learn the semantics of label names or metadata \cite{Zhang2020MinimallySC} to propagate document labels.

The above methods do not take aspect-specific opinion words into consideration. The semantic meaning captured by a $\lb$sentiment, aspect$\rb$ joint topic preserves more fine-grained information to imply the aspect of a sentence and thus can be used to improve the performance of aspect extraction.

\subsection{Joint Extraction of Aspect and Sentiment}\label{sec:rel_Joint}
Most previous studies that jointly perform aspect and sentiment extraction are LDA-based methods. \citet{Zhao2010JointlyMA} include aspect-specific opinion models along with aspect models in the generative process. \citet{Wang2015SentimentAspectEB} propose a restricted Boltzmann machine-based model that treats aspect and sentiment as heterogeneous hidden units. \citet{Xu2012TowardsJE} adapt LDA by introducing sentiment-related variables and integrating sentiment prior knowledge. All these methods rely on external resources such as part-of-speech (POS) tagging or opinion word lexicons. A more recent study that shares similar weakly-supervised setting with ours is W2VLDA \cite{GarcaPablos2018W2VLDAAU}. They apply Brown clustering \cite{Brown1992ClassBasedNM} to separate aspect-terms from opinion-terms and construct biased hyperparameters \textbf{$\alpha$} and \textbf{$\beta$} by embedding similarity. 
Despite the effectiveness of topic models, they suffer from the drawback of not imposing discriminative constraints among topics---topic-word distribution can largely overlap among different topics, allowing redundant topics to appear and making it hard to classify them. We empirically show the advances of our method by capturing discriminative joint topic representations in the embedding space.

\section{Problem Definition}

Our weakly-supervised aspect-based sentiment analysis task is defined as follows. 
The input is a training corpus $\mathcal{D}=\{d_1,d_2,\ldots,d_{|\mathcal{D}|}\}$ of text reviews from a certain domain (e.g., restaurant or laptop) \textit{without} any label for aspects or sentiment. A list of keywords $l_a$ for each aspect topic (denoted as $a \in A$) and $l_s$ for each sentiment polarity (denoted as $s \in S$) are provided by users as guidance. For each unseen review in the same domain, our model outputs a set of $\lb$$s$, $a$$\rb$ labels. 



\section{Model}
\begin{figure*}[t]
\centering
\includegraphics[width=0.98\linewidth]{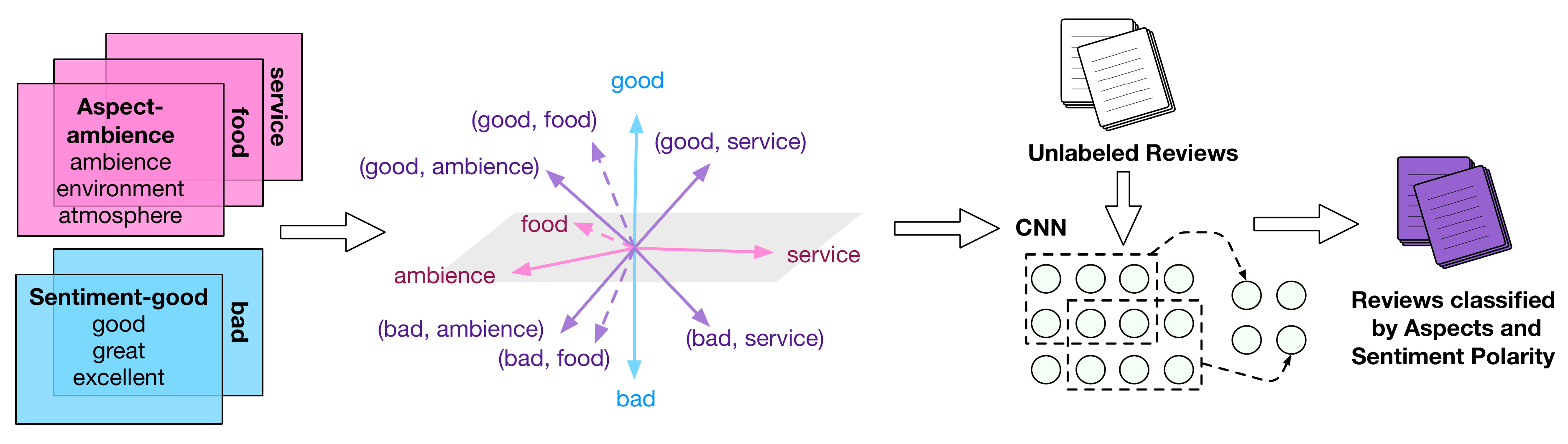}
\caption{Overview of our model \jasen. We first leverage the in-domain training corpus and user-given keywords to learn joint topic representation in the word embedding space. The marginal probability of keywords belonging to an aspect/sentiment can be summed up by the joint distribtution over $\lb$sentiment, aspect$\rb$ joint topics. Embedding-based prediction on unlabeled data are then leveraged by neural models for pre-training and self-training.
}
\label{fig:workflow}	
\vspace{-0.3cm}
\end{figure*}

\begin{figure}[t]
\centering
\includegraphics[width=1.0\linewidth]{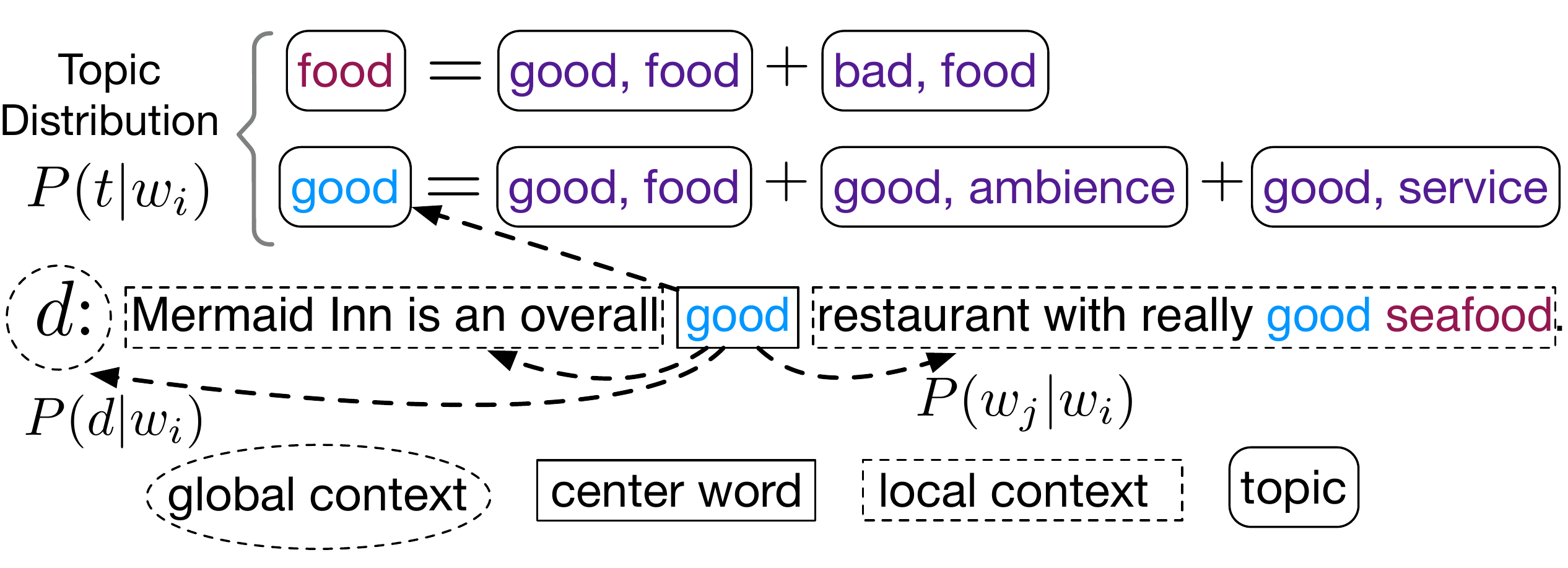}
\caption{Embedding training.
}
\label{fig:emb}	
\vspace{-0.3cm}
\end{figure}

Figure~\ref{fig:workflow} shows the workflow of \jasen.
Our goal is to generate a set of $\lb$sentiment, aspect$\rb$ predictions for each review. 

We first learn an embedding space to explicitly represent the semantics of the topics (including both pure aspect/sentiment and joint $\lb$sentiment, aspect$\rb$ ones) as embedding vectors, which are surrounded by the embeddings of the representative words of the topics. We also impose discriminative regularization on the embedding space to push different topics apart. To model the local sequential information which is crucial for sentiment analysis, we use CNN as the classifier by pre-training it on pseudo labels given by the cosine similarity between document embeddings and topic embeddings, and self-training it on unlabeled data to iteratively refine its parameters. Below we introduce the details of \jasen.




\subsection{Joint-Topic Representation Learning}
We learn the representations of words and topics on the in-domain corpus by following two principles: (1) distributional hypothesis~\cite{Sahlgren2008TheDH} and (2) topic distinctiveness. The first principle is achieved by an adaptation of the Skip-Gram model~\cite{Mikolov2013DistributedRO} through modeling both local and global contexts of words, and the second is achieved by a series of regularization objectives.
Fig.~\ref{fig:emb} provides the overview of our embedding learning objectives with an example.

\paragraph{Modeling Local and Global Contexts.}
We learn word embeddings based on the assumption that words with similar contexts have similar meanings, and define contexts to be a combination of location contexts~\cite{Mikolov2013DistributedRO} and global contexts~\cite{Meng2019SphericalTE, Liao2019CouplingGA,Meng2020DiscriminativeTM}. 
The local context of a word $w_i$ refers to other words whose distances are $h$ words or less from $w_i$. To maximize the probability of seeing the local context of a word $w_i$, we use the following objective: 
\begin{equation}
\label{eq:local}
\mathcal{L}_l = - \sum_{w_i} \sum_{0 < |j-i| \leq h} \log P(w_j|w_i),
\end{equation}
where $P(w_j|w_i) \propto \exp(\bs{v}_j^\top \bs{u}_i)$, and $\bs{u}_i, \bs{v}_j$ are the center and context word embeddings.

The global context~\cite{Meng2019SphericalTE, Liao2019CouplingGA} of a word $w_i$ refers to the document $d$ where a word appears, based on the motivation that similar documents contain similar-meaning words. We use the following objective for global context:
\begin{equation}
\label{eq:global}
\mathcal{L}_g = - \sum_{d\in \mathcal{D}} \sum_{w_i\in d} \log P(d|w_i),
\end{equation}
where $P(d|w_i) \propto \exp(\bs{d}^\top \bs{u}_i)$.

\paragraph{Regularizing Pure Aspect/Sentiment Topics.}
To endow the embedding space with discriminative power over the aspect/sentiment categories for better classification performance, we regularize the aspect topic embeddings $\bs{t}_a$ and sentiment topic embeddings $\bs{t}_s$ so that different topics are pushed apart.
For example, the word ``good'' in Fig.~\ref{fig:emb} is a keyword for the sentiment topic \textit{good}, and we aim to place $\bs{t}_{\text{good}}$ close to the word embedding of ``good'' in the embedding space while away from other topic embeddings (\ie, $\bs{t}_{\text{bad}}$). 
To achieve this, we maximize the probability of using each topic keyword to predict its representing topic:
\begin{align}
\label{eq:reg_a}
\mathcal{L}_{reg}^A &= -\sum_{a \in A} \sum_{w_i \in l_a} \log P(t_{a} | w_i),\\
\label{eq:reg_s}
\mathcal{L}_{reg}^S &= -\sum_{s \in S} \sum_{w_i \in l_s} \log P(t_{s} | w_i),
\end{align}
where $l_a$, $l_s$ are the keyword lists for aspect $a$ and sentiment $s$, respectively; $P(t|w_i) \propto \exp (\bs{u}_i^\top \bs{t})$. 
Eqs.~\eqref{eq:reg_a} and \eqref{eq:reg_s} empower the embedding space for classification purpose, that is, words can be ``classified'' into topics based on embedding similarity.
For good initializations of $\bs{t}_a$ and $\bs{t}_s$, we use the average word embedding of user-provided keywords for each aspect and sentiment.

\paragraph{Regularizing Joint $\lb$Sentiment, Aspect$\rb$ Topics.}
Now we examine the joint case, where $|S| \times |A|$ topics are regularized. 
We connect the learning of joint topic embeddings with pure aspect/sentiment topics by exploring the relationship between marginal distribution and joint distribution:
\begin{align}
\label{eq:marginal_a}
P(t_{a} | w_i) &= \sum_{s\in S} P \left( t_{\lb s,a \rb} \Big| w_i\right),\\
\label{eq:marginal_s}
P(t_{s} | w_i) &= \sum_{a\in A} P \left( t_{\lb s,a \rb} \Big| w_i\right).
\end{align}
As an example, 
Fig.~\ref{fig:emb} shows that the marginal probability of the keyword ``good'' belonging to the sentiment topic ``good'' is equal to the probability sum of it belonging to $\lb$good, food$\rb$, $\lb$good, ambience$\rb$ and $\lb$good, service$\rb$.

The objective for regularizing joint topics $\mathcal{L}_{joint}$ can be derived by replacing $P(t_a|w_i)$ in Eq.~\eqref{eq:reg_a} with Eq.~\eqref{eq:marginal_a} and $P(t_s|w_i)$ in Eq.~\eqref{eq:reg_s} with Eq.~\eqref{eq:marginal_s}. 

We also notice that general opinion words such as ``good'' (or pure aspect words such as ``seafood'') are equally irrelevant to the aspect (or sentiment) dimension, so we use a uniform distribution $\mathcal{U}$ to regularize their distribution over all the classes on the irrelevant dimension:
\begin{align}
\label{eq:cross}
\mathcal{L}_{cross}^A &= \sum_{s\in S} \sum_{w_i\in l_s} \text{KL} \left( \mathcal{U}, P(t_{a}|w_i) \right),\\
\mathcal{L}_{cross}^S &= \sum_{a\in A} \sum_{w_i\in l_a} \text{KL} \left( \mathcal{U}, P(t_{s}|w_i) \right).
\end{align}
Putting the above objectives altogether, our final embedding learning objective is:
\begin{equation}
\label{eq:emb}
\mathcal{L} = \mathcal{L}_l + \lambda_{g} \mathcal{L}_g + \lambda_{r} (\mathcal{L}_{reg} + \mathcal{L}_{joint} + \mathcal{L}_{cross}),
\end{equation}
where $\mathcal{L}_{reg} = \mathcal{L}_{reg}^A + \mathcal{L}_{reg}^S$, and the same for $\mathcal{L}_{joint}$ and $\mathcal{L}_{cross}$. For all the regularization terms, we treat them equally by using the same weight $\lambda_r$, which shows to be effective in practice.




\subsection{Training CNNs for Classification}

Word ordering information is crucial for sentiment analysis.
For example, ``Any movie is better than this one'' and ``this one is better than any movie'' convey opposite sentiment polarities but have the exactly same words.
The trained embedding space mainly captures word-level discriminative signals but is insufficient to model such sequential information. 
Therefore, we propose to train convolutional neural networks (CNNs) to generalize knowledge from the preliminary predictions given by the embedding space.
Specifically, we first pre-train CNNs with soft predictions given by the cosine similarity between document embeddings and topic embeddings, and then adopt a self-training strategy to further refine the CNNs using their high-confident predictions on unlabeled documents.

\paragraph{Neural Model Pre-training.}
For each unlabeled review, we can (1) derive one distribution over the joint topics by calculating the cosine similarity between the document representation $\bs{d}$ and $\bs{t}_{\lb s,a \rb}$, (2) derive separate distributions over sentiment and aspect variables using cosine similarity with marginal topics $\bs{t}_a$ and $\bs{t}_s$, or (3) combine (1) and (2) by adding the two sets of cosine scores.
We find empirically that the last method achieves the best result, \ie, the distribution of a test review $d$ over the aspect and sentiment categories is computed as:

{\small
\begin{align}
\label{eq:pre_score_a}
P(a|d) &\propto \exp \left( T \cdot \left(\cos(\bs{t}_a, \bs{d}) + \frac{\sum_{s\in S}\cos(\bs{t}_{\lb s,a \rb}, \bs{d})}{|S|} \right) \right), \\
\label{eq:pre_score_s}
P(s|d) &\propto \exp \left(T \cdot \left(\cos(\bs{t}_s, \bs{d}) + \frac{\sum_{a\in A}\cos(\bs{t}_{\lb s,a \rb}, \bs{d})}{|A|} \right) \right),
\end{align}} 
\noindent 
where $\bs{d}$ is obtained by averaging the word embeddings in $d$, and $T$ is the temperature to control how greedy we want to learn from the embedding-based prediction.

We train two CNN models separately for aspect and sentiment classification by learning from the two distributions in Eqs.~\eqref{eq:pre_score_a} and \eqref{eq:pre_score_s}. We leverage the knowledge distillation objective~\cite{Hinton2015DistillingTK} to minimize the cross entropy between the embedding-based prediction $p_d$ and the output prediction $q_d$ of the CNNs:
\begin{equation}
H(p_d,q_d) = - \sum_{t} P(t|d) \log Q(t|d).
\end{equation}

\paragraph{Neural Model Refinement.}
The pre-trained CNNs only copy the knowledge from the embedding space. 
To generalize their current knowledge to the unlabeled corpus, we adopt a self-training technique to bootstrap the CNNs. 
The idea of self-training is to use the model's high-confident predictions on unlabeled samples to refine itself.
Specifically, we compute a target score~\cite{Xie2016UnsupervisedDE} for each unlabeled document based on the predictions of the current model by enhancing high-confident predictions via a squaring operation:
$$\text{target}(P(a|d)) = \frac{P(a|d)^2/f_a}{\sum_{a'\in A}P(a'|d)^2/f_{a'}},$$
where $f_{a}=\sum_{d\in \mathcal{D}} P(a|d)$. Since self-training updates the target scores at each epoch, the model is gradually refined by its most recent high-confident predictions. The self-training process is terminated when no more samples change label assignments after the target scores are updated. The resulting model can be used to classify any unseen reviews.

\section{Evaluation}


We conduct a series of quantitative and qualitative evaluation on benchmark datasets to demonstrate the effectiveness of our model.

\begin{table}
\centering
\scalebox{0.8}{
\begin{tabular}{|c|c|c|}
\hline
Dataset & \#Training reviews & \#Test reviews\\
\hline
\textbf{Restaurant} & 17,027 & 643\\
\hline
\textbf{Laptop} & 14,683 & 307\\
\hline
\end{tabular}
}\caption{Dataset Statistics.}\label{tab:dataset}
\end{table}

\begin{table}
\centering

\scalebox{0.8}{
\begin{tabular}{|c|c|c|}
\hline
Dataset & Aspect & Keywords\\
\hline
\multirow{5}{*}{\makecell{ \textbf{Restaurant}}} & Location
& \makecell{street block river avenue}\\
\cline{2-3}
& Drinks 
& \makecell{beverage wines cocktail sake}\\
\cline{2-3}
& Food
& \makecell{spicy sushi pizza taste}\\
\cline{2-3}
& Ambience
& \makecell{atmosphere room \\ seating environment}\\
\cline{2-3}
& Service
& \makecell{tips manager waitress servers}\\
\hline
\multirow{8}{*}{\makecell{\textbf{Laptop}}} 
& Support
& \makecell{service warranty\\ coverage replace}\\
\cline{2-3}
& OS
& \makecell{windows ios mac system}\\
\cline{2-3}
& Display
& \makecell{screen led monitor resolution}\\
\cline{2-3}
& Battery
& \makecell{life charge last power}\\
\cline{2-3}
& Company
& \makecell{hp toshiba dell lenovo}\\
\cline{2-3}
& Mouse
& \makecell{touch track button pad}\\
\cline{2-3}
& Software
& \makecell{programs apps itunes photoshop}\\
\cline{2-3}
& Keyboard
& \makecell{key space type keys} \\
\hline
\end{tabular}

}\caption{Keywords of each aspect.}\label{tab:keyword}
\vspace{-0.3cm}
\end{table}

\subsection{Experimental Setup}
\smallskip
\noindent \textbf{Datasets:} The following two datasets are used for evaluation:
\begin{itemize}[leftmargin=*]
\parskip -0.2ex
    \item \textbf{Restaurant}: For in-domain training corpus, we collect 17,027 unlabeled reviews from \textit{Yelp Dataset Challenge}\footnote{https://www.yelp.com/dataset/challenge}. For evaluation, we use the benchmark dataset in the restaurant domain in SemEval-2016 \cite{pontiki-etal-2016-semeval} and SemEval-2015 \cite{Pontiki2015SemEval2015T1}, where each sentence is labeled with aspect and sentiment polarity. We remove sentences with multiple labels or with a \textit{neutral} sentiment polarity to simplify the problem (otherwise a set of keywords can be added to describe it).
    \item \textbf{Laptop}: We leverage 14,683 unlabeled Amazon reviews under the laptop category collected by~\cite{He2016UpsAD} as in-domain training corpus. We also use the benchmark dataset in the laptop domain in SemEval-2016 and SemEval-2015 for evaluation. 
    Detailed statistics of both datasets are listed in Table \ref{tab:dataset}, and the aspects along with their keywords are in Table \ref{tab:keyword}.
\end{itemize}

\vspace{-0.2cm}
\smallskip
\noindent \textbf{Preprocessing and Hyperparameter Setting.}
To preprocess the training corpus $D$, we use the word tokenizer provided by \textit{NLTK}\footnote{https://www.nltk.org/}. We also use a phrase mining tool, AutoPhrase \cite{Shang2017AutomatedPM}, to extract meaningful phrases such as ``great wine'' and ``numeric keypad'' such that they can capture complicated semantics in a single text unit. We use the benchmark validation set to fine-tune the hyperparameters: embedding dimension = $100$, local context window size $h=5$, $\lambda_g = 2.5$, $\lambda_r = 1$, training epoch = $5$. For neural model pre-training, we set $T = 20$. A CNN model is trained for each sub-task: aspect extraction and sentiment classification. Each model uses 20 feature maps for filters with window sizes of 2, 3, and 4. SGD is used with $1e-3$ as the learning rate in both pre-training and self-training and the batch size is set to 16.

\begin{table*}[ht]
	\centering
	\small
	
	\scalebox{0.9}{
		\begin{tabular}{c|cccc|cccc}
			\toprule
			\multirow{2}{*}{Methods} &
			\multicolumn{4}{c|}{\textbf{Restaurant}} & \multicolumn{4}{c}{\textbf{Laptop}} \\
			& Accuracy & Precision & Recall & macro-F1
			& Accuracy & Precision & Recall & macro-F1 \\
			\midrule
			CosSim 
			& 61.43 & 50.12 & 50.26 & 42.31 
			& 53.84 & 58.79 & 54.64 & 52.18\\
			ABAE\cite{He2017AnUN} 
			& 67.34 & 46.63 & 50.79 & 45.31
			& 59.84 & 59.96 & 59.60 & 56.21\\
			CAt\cite{Tulkens2020EmbarrassinglySU}
			& 66.30 & 49.20 & 50.61 & 46.18   
			& 57.95 & 65.23 & 59.91 & 58.64 \\
			W2VLDA\cite{GarcaPablos2018W2VLDAAU}
			& 70.75 & 58.82 & 57.44 & 51.40  
			& 64.94 & 67.78 & 65.79 & 63.44 \\
			BERT\cite{Devlin2019BERTPO}
			& 72.98 & 58.20 & \textbf{74.63} & 55.72   
			& 67.52 & 68.26 & 67.29 & 65.45 \\
			\jasen w/o joint & 81.03 & 61.66 & 65.91 & 61.43& 69.71 & 69.13 & 70.65 & 67.49 \\
            \jasen w/o self train
			& 82.90 & 63.15 & 72.51 & 64.94 
			& 70.36 & 68.77 & 70.91 & 68.79  \\
			\jasen
			& \textbf{83.83} & \textbf{64.73} & 72.95 & \textbf{66.28} 
			& \textbf{71.01} & \textbf{69.55} & \textbf{71.31} & \textbf{69.69}  \\
			\bottomrule
		\end{tabular}
		
	}\caption{Quantitative evaluation on aspect identification (\%).}\label{tab:aspect}
\end{table*}

\begin{table*}[ht]
	\centering
	\small
	\scalebox{1.0}{
		\begin{tabular}{c|cccc|cccc}
			\toprule
			\multirow{2}{*}{Methods} &
			\multicolumn{4}{c|}{\textbf{Restaurant}} & \multicolumn{4}{c}{\textbf{Laptop}} \\
			& Accuracy & Precision & Recall & macro-F1
			& Accuracy & Precision & Recall & macro-F1 \\
			\midrule
			CosSim 
			& 70.14 & 74.72 & 61.26 & 59.89
			& 68.73 & 69.91 & 68.95 & 68.41\\
			W2VLDA
			& 74.32 & 75.66 & 70.52 & 67.23  
			& 71.06 & 71.62 & 71.37 & 71.22 \\
			BERT
			& 77.48 & 77.62 & 73.95 & 73.82   
			& 69.71 & 70.10 & 70.26 & 70.08 \\
            \jasen w/o joint & 78.07 & 80.60 & 72.40 & 73.71 & 72.31 & 72.34 & 72.25 & 72.26\\
            \jasen w/o self train
			& 79.16 & 81.31 & 73.94 & 75.34 
			& 73.29 & 73.69 & 73.42 & 73.24  \\
			\jasen
			& \textbf{81.96} & \textbf{82.85} & \textbf{78.11} & \textbf{79.44} 
			& \textbf{74.59} & \textbf{74.69} & \textbf{74.65} & \textbf{74.59}  \\
			\bottomrule
		\end{tabular}
		
	}\caption{Quantitative evaluation on sentiment polarity classification (\%).}\label{tab:senti}
\end{table*}

\subsection{Quantitative Evaluation}
We conduct quantitative evaluation on both aspect extraction and sentiment polarity classification.

\smallskip
\noindent \textbf{Compared Methods.}
Our model is compared with several previous studies. A few of them are specifically designed for aspect extraction but do not perform well on sentiment classification. So we only report their results on aspect extraction. For fair comparison, we use the same training corpus and test set for each baseline method. For weakly-supervised methods, they are fed with the same keyword list as ours.
\begin{itemize}[leftmargin=*]
\parskip -0.2ex
    \item \textbf{CosSim}: The topic representation is averaged by the embedding of seed words trained by Word2Vec on training corpus. Cosine similarity is computed between a test sample and the topics to classify the sentence.
    \item \textbf{ABAE} \cite{He2017AnUN}: An attention-based model to unsupervisedly extract aspects. An autoencoder is trained to reconstruct sentences through aspect embeddings. The learned topics need to be manually mapped to aspects.
    \item \textbf{CAt} \cite{Tulkens2020EmbarrassinglySU}: A recent method for unsupervised aspect extraction. A single head attention is calculated by a Radio Basis Function kernel to be the sentence summary.
    \item \textbf{W2VLDA} \cite{GarcaPablos2018W2VLDAAU}: A state-of-the-art topic modeling based method that leverages keywords for each aspect/sentiment to jointly do aspect/sentiment classification.
    \item \textbf{BERT} \cite{Devlin2019BERTPO}: A recent proposed deep language model. We utilize the pre-trained BERT (12-layer, 768 dimension, uncased) and implement a simple weakly-supervised method that fine-tunes the model by providing pseudo labels for sentences containing keywords from a given aspect/sentiment.
    \item \textbf{\jasen w/o joint}: An ablation of our proposed model. Neural model is pre-trained on separate topic embedding for each sentiment and aspect.
    \item \textbf{\jasen w/o self train}: An ablation of our proposed model without self-training process.
\end{itemize}

\begin{table*}[ht]
\centering
	\small
	
	\scalebox{0.9}{
	\begin{tabular}{|c|c|c|c|c|c|c|}
	\hline
	& Ambience & Service & Food & Support & Keyboard & Battery \\
	\hline
	Good 
	& \makecell{cozy,\\intimate,\\comfortable,\\loungy,\\ great music}
	& \makecell{professional,\\ polite,\\knowledgable,\\informative,\\helpful}
	& \makecell{huge portion,\\flavourful,\\super fresh,\\ husband loves,\\authentic italian}
	& \makecell{accidental damage\\protection, 
	accidental\\damage warranty, 
	generous,\\guarantee,\\commitment}
	& \makecell{tactile feedback,\\tactile feel,\\ classic,\\nicely spaced,\\chiclet style}
	& \makecell{lasts long,\\charges quickly,\\high performance,\\lasting,\\great power}
	\\
	\hline
	Bad
	& \makecell{cramped,\\unbearable,\\uncomfortable,\\dreary, chaos}
	& \makecell{inattentive,\\ignoring,\\extremely rude,\\condescending,\\inexperienced}
	& \makecell{microwaved,\\flavorless,\\vomit,\\frozen food,\\undercooked}
	& \makecell{completely useless,\\denied,\\refused,\\blamed,\\apologize}
	& \makecell{large hands,\\shallow,\\cramped,\\wrong key,\\typos}
	& \makecell{completely dead,\\drained,\\ discharge,\\unplugged,\\torture}
	\\
	\hline
	\end{tabular}}
		\caption{Keywords retrieved by joint topic representations.}\label{tab:qual}
\end{table*}

\smallskip
\noindent \textbf{Aspect Extraction.}~
We report the results of aspect extraction of our model and all the baselines in Table \ref{tab:aspect}. We use four metrics for evaluation: Accuracy, Precision, Recall and macro-F1 score. 
We observe that weakly-supervised methods tend to have a better performance than unsupervised ones, suggesting that using keywords to enrich the semantics of labels is a promising direction to increase classification performance.
As shown in Table \ref{tab:aspect}, our model, even without self-training, outperforms baseline methods on most of the metrics by a large margin, indicating that \jasen obtains substantial benefits from learning the semantics of fine-grained joint topics, and self-training boosts the performance further. 
We observe that \jasen can deal with cases where the target of the sentence is not explicitly mentioned. 
For example, \jasen correctly labels ``It's to die for!'' as (good, food). Although nothing mentioned is related to food, ``to die for'' appears in other sentences addressing the tastiness of food, thus is captured by the joint topic of (good, food).

\smallskip
\noindent \textbf{Sentiment Polarity Classification.}~
We compare \jasen against baseline methods on sentiment classification and show the results in Table \ref{tab:senti}. Since some methods are designed for aspect extraction and do not perform well on sentiment classification, we do not report their results. 
As shown in the table, \jasen outperforms all the baselines on both datasets.
We also observe that methods only leveraging local contexts do not perform well compared to methods that leverage both global and local contexts on \textbf{Laptop} dataset. Since ``good'' and ``bad'' are a pair of antonyms, they can have very similar collocations, so models purely capturing local contexts do not distinguish them well.

\subsection{Qualitative Evaluation}

To evaluate the quality of the joint topic representation, we retrieve their representative terms by ranking the embedding cosine similarity between words and
each joint topic vector. For brevity, we randomly sample 3 aspects from each dataset and pair them up with the two sentiment polarity to form 12 joint topics. We list their top terms in Table \ref{tab:qual}.
Results show that the representative terms form coherent and meaningful topics, and they are not restricted to be adjectives, such as ``vomit'' in (bad, food) and ``commitment'' in (good, support). Another interesting observation is that ``cramped'' appears in both (bad, ambience) in restaurant domain and (bad, keyboard) in laptop domain, suggesting that \jasen captures different meanings of words based on in-domain corpus.

\begin{figure}[ht]
\centering
\includegraphics[width=1\linewidth]{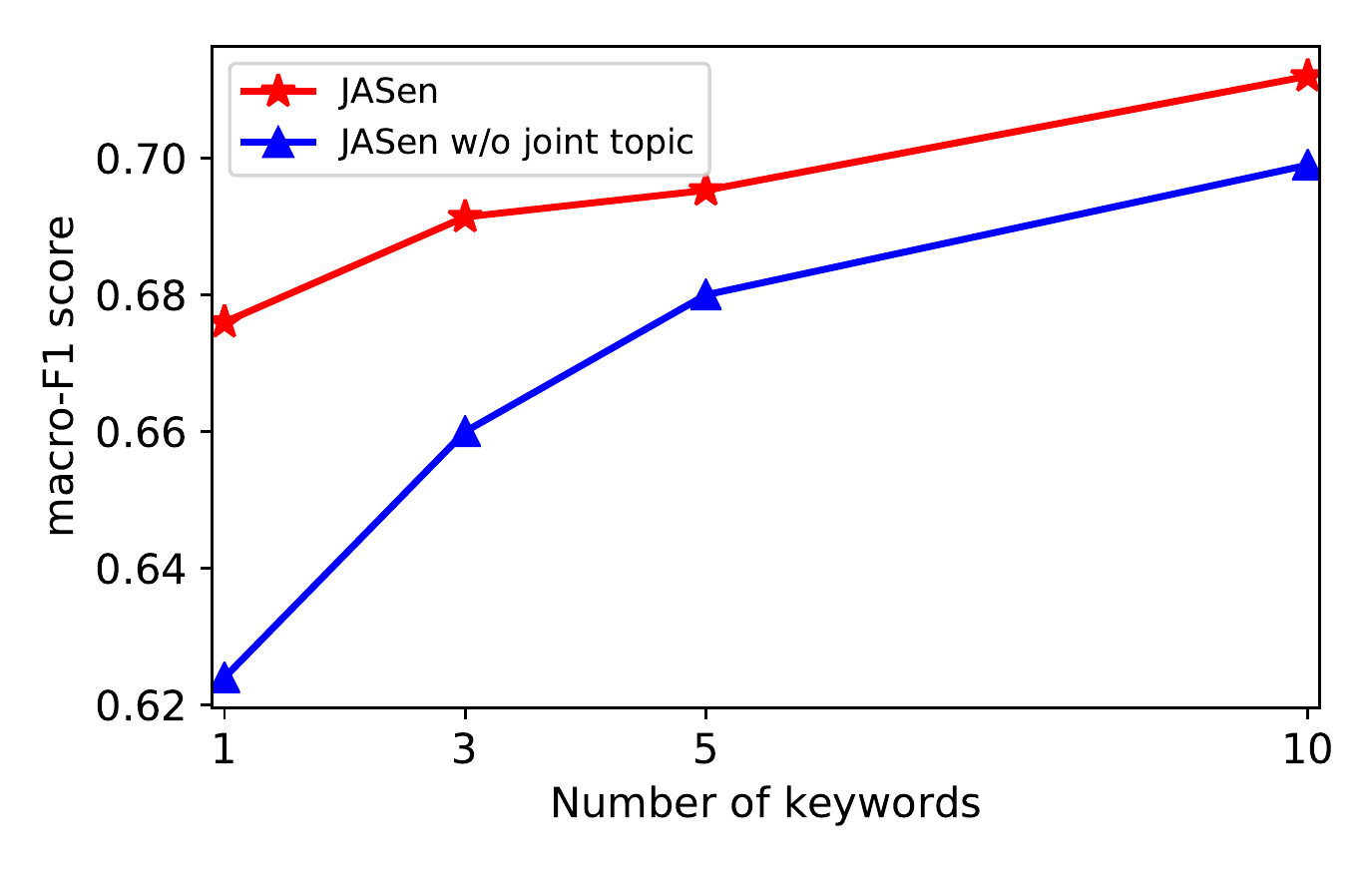}
\caption{Macro-F1 score vs. keyword number.}
\label{fig:kw}
\end{figure}

\subsection{Effect of Number of Keywords}
We study the effect of the number of keywords. In Figure \ref{fig:kw} we show the macro-F1 score of aspect extraction using different number of keywords for each aspect on \textbf{Laptop} dataset. The trend clearly shows the model performance increases when more keywords are provided. Moreover, when only one keyword is provided (only the label name), \jasen still has a stable performance and a large gap over the ablation without learning joint topic embedding, implying that learning joint topic semantics is especially powerful in low resource setting.

\begin{figure}[ht]
\centering
\includegraphics[width=1\linewidth]{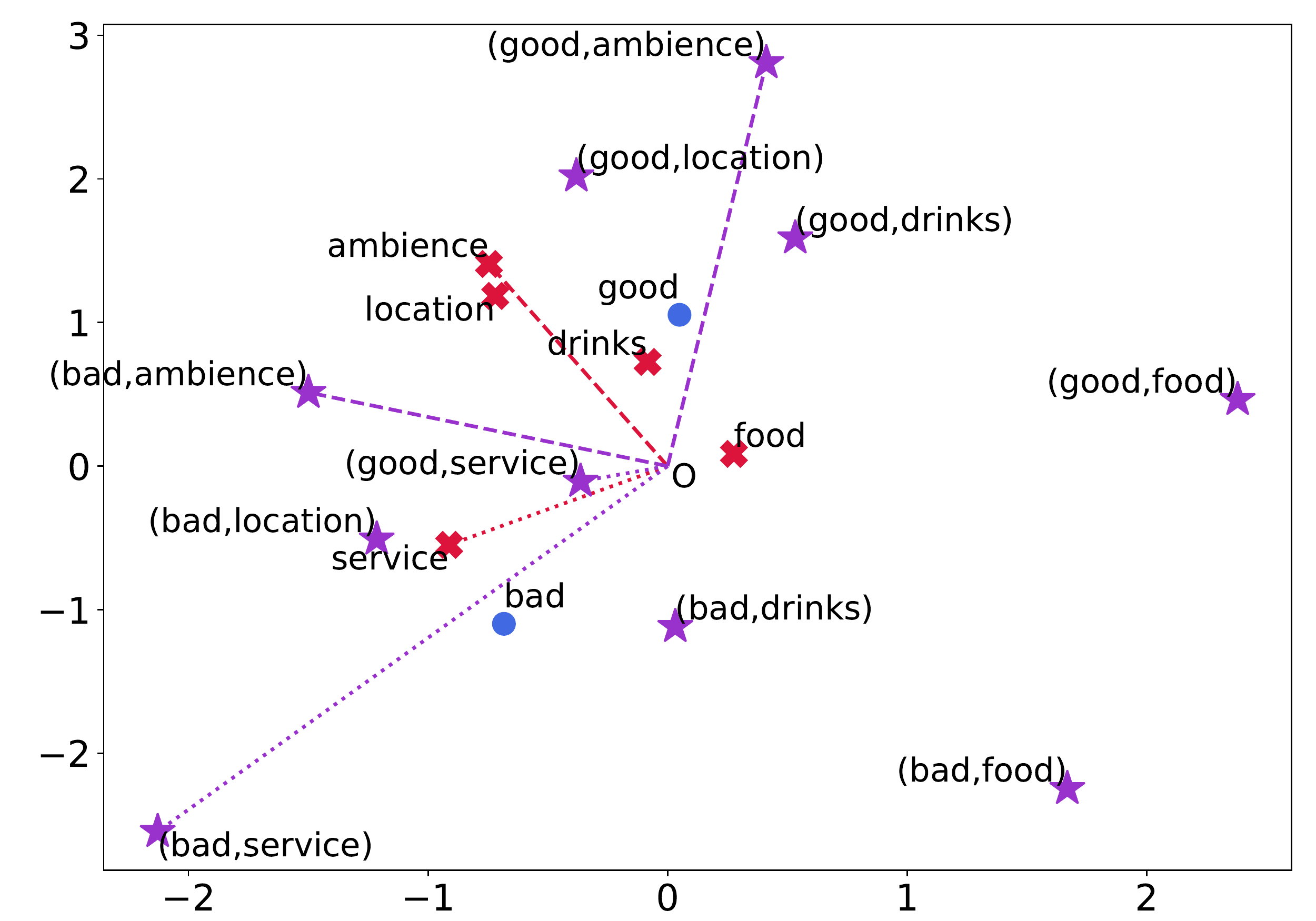}
\caption{Visualization of joint topics (purple stars), aspect topics (red crosses) and sentiment topics (blue dots) in the embedding space.}
\label{fig:vis}
\end{figure}

\subsection{Joint Topic Representation Visualization}
To understand how the joint topics are distributed in the embedding space, along with the aspect and sentiment topics, we use PCA~\cite{Jolliffe2011PrincipalCA} for dimension reduction to visualize topic embedding trained on the \textbf{Restaurant} corpus in Fig.\ \ref{fig:vis}. An interesting observation is that, some aspect topics (e.g., \textit{ambience}) lie approximately in the middle of their joint topics (``\textit{good, ambience}'' and ``\textit{bad, ambience}''), showing that our embedding learning objective understands the joint topics as decomposition of their ``marginal'' topics, which fits with our goal to learn fine-grained topics.

\begin{table*}[ht]
\centering
	\small
	\scalebox{0.9}{
	\begin{tabular}{|c|c|c|c|}
	\hline
	 Review & Ground Truth & \makecell{Output of\\ Full Model} & \makecell{Output of Model\\ w/o joint embedding} \\
	\hline
	\makecell{
	The wait staff is very freindly, they make it\\
	feel like you're eating in a freindly little\\ european town.}
	& (good, service) & (good, ambience) & (good, location)\\
	\hline
	\makecell{
	The outdoor atmosphere of sitting on the sidewalk\\ watching the world go by 50 feet away on 6th \\
	avenue on a cool evening was wonderful. 
	}
	& (good, location) & (good, ambience) & (good, ambience)\\
	\hline
	\makecell{
	It's simply the best meal in NYC.}
	& (good, food) & (good, food) & (good, location)\\
	\hline
	\makecell{
	You can get a table without a reservation if you\\ get there early I they don't make you by bottles.}
	& (good, service) & (good, service) & (bad, service)\\
	\hline
	\makecell{
	The sauce tasted more like Chinese fast food than\\ decent Korean.} 
	& (bad, food) & (good, food) & (bad, food)\\
	\hline
	\makecell{
	My wife had barely\#\#\#touched that mess of a dish.}
	& (bad, food) & (bad, food) & (good, food)\\
	\hline
	\makecell{
	This is undoubtedly my favorite modern Japanese\\ brasserie (that don’t serve sushi), and in my\\ opinion, one of the most romantic restaurants in\\ the city!}
	& (good, ambience) & (good, food) & (good, location)
	\\
	\hline
	\makecell{
	We took advanatage of the half price sushi deal\\
	on saturday so it was well worth it.}
	& (good, food) & (good, food) & (bad, food) \\
	\hline
	\makecell{ 
	If you don't like it, I don't know what to tell you.}
	& (good, food) & (bad, food) & (bad, service)\\
	\hline

	
	\end{tabular}}
		\caption{Comparison of predictions on sample \textbf{Restaurant} reviews between our full model and model pre-trained w/o joint topic embedding. }\label{tab:cas_rest}
\end{table*}

\begin{table*}[h]
\centering
	\small
	\scalebox{0.9}{
	\begin{tabular}{|c|c|c|c|}
	\hline
	 Review & Ground Truth & \makecell{Output of\\ Full Model} & \makecell{Output of Model\\ w/o joint embedding} \\
	\hline
	\makecell{
	NO junkware!!}
	& (good, software) & (good, software) & (good, display) \\
	\hline
	\makecell{ 
	 I definitely will buy a Mac again if and when this\\ computer ever fails.}
	& (good, company) & (good, os) & (good, os) \\
	\hline
	\makecell{ 
	I don't have the inclination or time to devote to a\\ companies tech support, search functions, or hold\\ times.....dropped the HP and never looked back.}
	& (bad, company) & (bad, support) & (bad, company)\\
	\hline
	\makecell{I find myself adjusting it regularly.}
	& (bad, display) & (bad, display) & (bad, mouse)\\
	\hline
	\makecell{ 
	I thought learning the Mac OS would be hard, but\\
	it is easily picked up if you are familiar with a PC.}
	& (good, os) & (good, os) & (bad, os) \\
	\hline
	\makecell{
	They told me to reprogram the computer, and I \\
	didn't need  to do that, and I lost pictures\\
	of my oldests first 2 years of her life.}
	& (bad, support) & (bad, support) & (good, support)\\
	\hline
	\makecell{But, hey, it's an Apple.}
	& (good, company) & (bad, company) & (good, company)\\
	\hline
	\makecell{ 
	I'm no power\#\#\#user, but I have had no\\
	learning\#\#\#curve with the MAC and I don't do\\
	anything geeky enough forcing me to learn the OS.}
	& (good, os) & (good, os) & (bad, os)\\
	\hline
	\makecell{
	The battery lasted 12 months, then pffft.....gone.}
	& (bad, battery) & (bad, battery) & (good, battery)\\
	\hline
	\end{tabular}}
		\caption{Comparison of predictions on sample \textbf{Laptop} reviews between our full model and model pre-trained w/o joint topic embedding.}\label{tab:cas_laptop}
\end{table*}

\subsection{Case Studies}
We list several test samples along with their ground truth and model predictions in Table \ref{tab:cas_rest} and Table \ref{tab:cas_laptop}.
Some conflicting cases between ours and the ground truth are rather ambiguous. For example, the ground truth of the second example in Table \ref{tab:cas_rest} is (good, location), but we still think given that the review mentions ``the outdoor atmosphere'' and uses terms like ``sitting on the sidewalk'' and ``cool evening'', it is more relevant to ambience than location, as is output by our full model. 
The gold aspect label for the second and the third reviews in Table \ref{tab:cas_laptop} are both ``company'', but apparently these two sentences are talking about two different aspects: the product itself and the service of the company. Though the output of our model, ``os'' and ``support'' for these two sentences may not be the most precise prediction, at least our model treats them as two different aspects.




\section{Conclusion}
In this paper we propose to enhance weakly-supervised aspect-based sentiment analysis by learning the representation of $\lb$sentiment, aspect$\rb$ joint topic in the embedding space to capture more fine-grained information. 
We introduce an embedding learning objective that leverages user-given keywords for each aspect/sentiment and models their distribution over the joint topics. 
The embedding-based predictions are then used for pre-training neural models, which are further refined via self-training on unlabeled corpus.
Experiments show that our method learns high-quality joint topics and outperforms previous studies substantially. 

In the future, we plan to adapt our methods to more general applications that are not restricted to the field of sentiment analysis, such as doing multiple-dimension classification (e.g., topic, location) on general text corpus. Another promising direction is to leverage taxonomy construction algorithms \cite{Huang2020CoRelST} to capture more fine-grained aspects, such as ``smell'' and ``taste'' for ``food''.

\section*{Acknowledgments}
Research was sponsored in part by US DARPA KAIROS Program No. FA8750-19-2-1004 and SocialSim Program No.  W911NF-17-C-0099, National Science Foundation IIS 16-18481, IIS 17-04532, and IIS 17-41317, and DTRA HDTRA11810026. Any opinions, findings, and conclusions or recommendations expressed herein are those of the authors and should not be interpreted as necessarily representing the views, either expressed or implied, of DARPA or the U.S. Government. The U.S. Government is authorized to reproduce and distribute reprints for government purposes notwithstanding any copyright annotation hereon. 
We thank anonymous reviewers for valuable and insightful feedback.

\bibliographystyle{acl_natbib}
\bibliography{emnlp2020}




\end{document}